# MatSciBERT: A Materials Domain Language Model for Text Mining and Information Extraction

Tanishq Gupta[1], Mohd Zaki[2], N. M. Anoop Krishnan[2,3,*], Mausam[3,4,*]

[1]*Deparment of Mathematics, Indian Institute of Technology Delhi, Hauz Khas, New Delhi, India 110016*

[2]*Deparment of Civil Engineering, Indian Institute of Technology Delhi, Hauz Khas, New Delhi, India 110016*

[3]*School of Artificial Intelligence, Indian Institute of Technology Delhi, Hauz Khas, New Delhi 110016*

[4]*Deparment of Computer Science and Engineering, Indian Institute of Technology Delhi, Hauz Khas, New Delhi, India 110016*

[*]Corresponding authors: N. M. A. Krishnan (krishnan@iitd.ac.in), Mausam (mausam@iitd.ac.in)

**Abstract**
An overwhelmingly large amount of knowledge in the materials domain is generated and stored as text published in peer-reviewed scientific literature. Recent developments in natural language processing, such as bidirectional encoder representations from transformers (BERT) models, provide promising tools to extract information from these texts. However, direct application of these models in the materials domain may yield suboptimal results as the models themselves may not be trained on notations and jargon that are specific to the domain. Here, we present a materials-aware language model, namely, MatSciBERT, which is trained on a large corpus of scientific literature published in the materials domain. We further evaluate the performance of MatSciBERT on three downstream tasks, namely, abstract classification, named entity recognition, and relation extraction, on different materials datasets. We show that MatSciBERT outperforms SciBERT, a language model trained on science corpus, on all the tasks. Further, we discuss some of the applications of MatSciBERT in the materials domain for extracting information, which can, in turn, contribute to materials discovery or optimization. Finally, to make the work accessible to the larger materials community, we make the pretrained and finetuned weights and the models of MatSciBERT freely accessible.

## 1 Introduction
The naming of human civilization after the materials used in those periods signifies their importance in our daily lives. Historically starting from the stone, bronze and iron ages, we have now come to the times when we are using a variety of alloys like magnesium[1] and aluminum[2] alloys in aerospace and automobiles, titanium alloys in biocompatible implants[3], glasses for optical and communication devices[4], and concrete for construction activities[5]. Despite technological advancement in experimental and computation domains, discovering new materials and bringing them to market is still a time-consuming process that may span decades[6,7]. To accelerate this process, we need to understand the properties of existing elements and their compounds which are the building block of materials[8–13]. Textbooks, scientific publications, reports, handbooks, websites etc., serve as a large data repository that can be mined for obtaining the already existing desired information[14,15]. However, it is a humanly impossible task to go through all the available literature and extract relevant information. Further, advancements in machine learning (ML) and natural language processing (NLP) have enabled researchers to automate the information extraction from the text. Although it is a challenging task, since most of the information in scientific journals is unstructured, for example, experimental procedures are reported in the form of paragraphs and image captions having findings of experiments observed using graphs and different microscopy and diffraction techniques.

In the material science domain, researchers have used NLP tools to automate database creation for ML applications. One such example is ChemDataExtractor[16], an NLP pipeline used to create databases of battery materials[17], Curie and Néel temperatures of magnetic materials[18], and inorganic material synthesis routes[19], thus demonstrating NLP applications' ability to understand the chemical and material science entities from the scientific text. In other related works, researchers have used NLP to collect the composition and dissolution rate of calcium aluminosilicate glassy materials from tables published in research articles and, hence, predict the property using ML algorithms[20]. Similarly, using the table extraction tools, the researchers have used ML on a dataset created through NLP assisted automated extraction of zeolite synthesis routes to synthesize germanium containing zeolites. In glass science, researchers have used latent-Dirichlet allocation to classify the literature into 15 broad categories. Further, using caption cluster plots, researchers have made it possible to find papers having specific elements, results from different experimental techniques and are related to broad categories of glass science[15]. The use of NLP to extract process and testing parameters of oxide glasses followed by including the extracted parameters as input parameters has improved the prediction of the Vickers hardness[21]. This increased use of NLP and ML in the material science domain implies the need for a universal NLP tool that can understand the domain-specific entities and provide convenient solutions for such automated extractions.

A comprehensive review by Olivetti et al. (2019) describes several ways in which NLP can benefit the material science community[22]. Providing insights into chemical parsing tools like OSCAR4[23] capable of identifying entities and chemicals from text, Artificial Chemist[24], which takes the input of precursor information and generates synthetic routes to manufacture optoelectronic semiconductors with targeted band gaps, robotic system for making thin films to produce cleaner and sustainable energy solutions[25], and identification of more than 80 million material science domain-specific named entities, researches have prompted the accelerated discovery of novel materials for different applications through the combination of ML and NLP techniques.

For using text in NLP applications, non-neural methods are based on n-grams such as Brown et al. (1992)[27], structural learning framework by Ando and Zhang (2005)[28], or structural correspondence learning by Blitzer et al. (2006)[29], but these are no longer the state of the art. Neural pre-trained embeddings like word2vec[30,31] and GloVe[32] are quite popular, but lack material science knowledge, and do not produce contextual embeddings -- a word embedding in the context of the accompanying text. Recent progress in NLP has led to the development of a novel computational paradigm in which a large, pre-trained language model (LM) is fine-tuned for domain specific tasks. Research has consistently shown that this pretrain-finetune paradigm leads to the best overall task performance. Statistically, LMs are probability distributions for a sequence of words such that for a given set of words, it assigns probability to each word[26]. Recently, due to availability of large amounts of text and high computing power, researchers are able to pre-train these large neural language models. For example, Bidirectional Encoder Representations from Transformers (BERT)[33] is trained on BookCorpus[34] and English Wikipedia, resulting in a state-of-the-art performance on multiple NLP tasks like question answering and entity recognition, to name a few.

Although researchers have shown the domain adaptation capability of word2vec and BERT in the field of biological sciences as BioWordVec[35] and BioBERT[36], other domain-specific BERTs like SciBERT[37] trained on scientific and biomedical corpus[38], clinicalBERT[39] trained on 2 million clinical notes in MIMIC-III v1.4 database[40], mBERT[41] for multilingual machine translations tasks, patentBERT[42] for patent classification and FinBERT for financial tasks[43], there is a lack of material science aware LM which can accelerate the research in the field by further adapting to downstream tasks. This has been indeed cited as a major challenge in several previous works[14,22]. Therefore, in this work, we train material science domain-specific BERT and achieve state of the art results on domain-specific tasks as listed below, details of which are described in the Results *and discussion* section of the paper.

a. NER on SOFC, SOFC Slot dataset by Friedrich et al. (2020)[44] and Matscholar dataset by Weston et al. (2019)[14]
b. Glass vs Non-Glass classification[15]
c. Relation Classification on MSPT corpus[45]

The present work, thus, bridges the gap in the availability of a materials domain language mode allowing researchers to automate information extraction, knowledge graph completion and hence accelerate the discovery of novel materials.

## 2 Methodology

Figure 1 shows the graphical summary of the methodology adopted in this work encompassing creating the material science corpus, training the MatSciBERT, and evaluating on different downstream tasks.

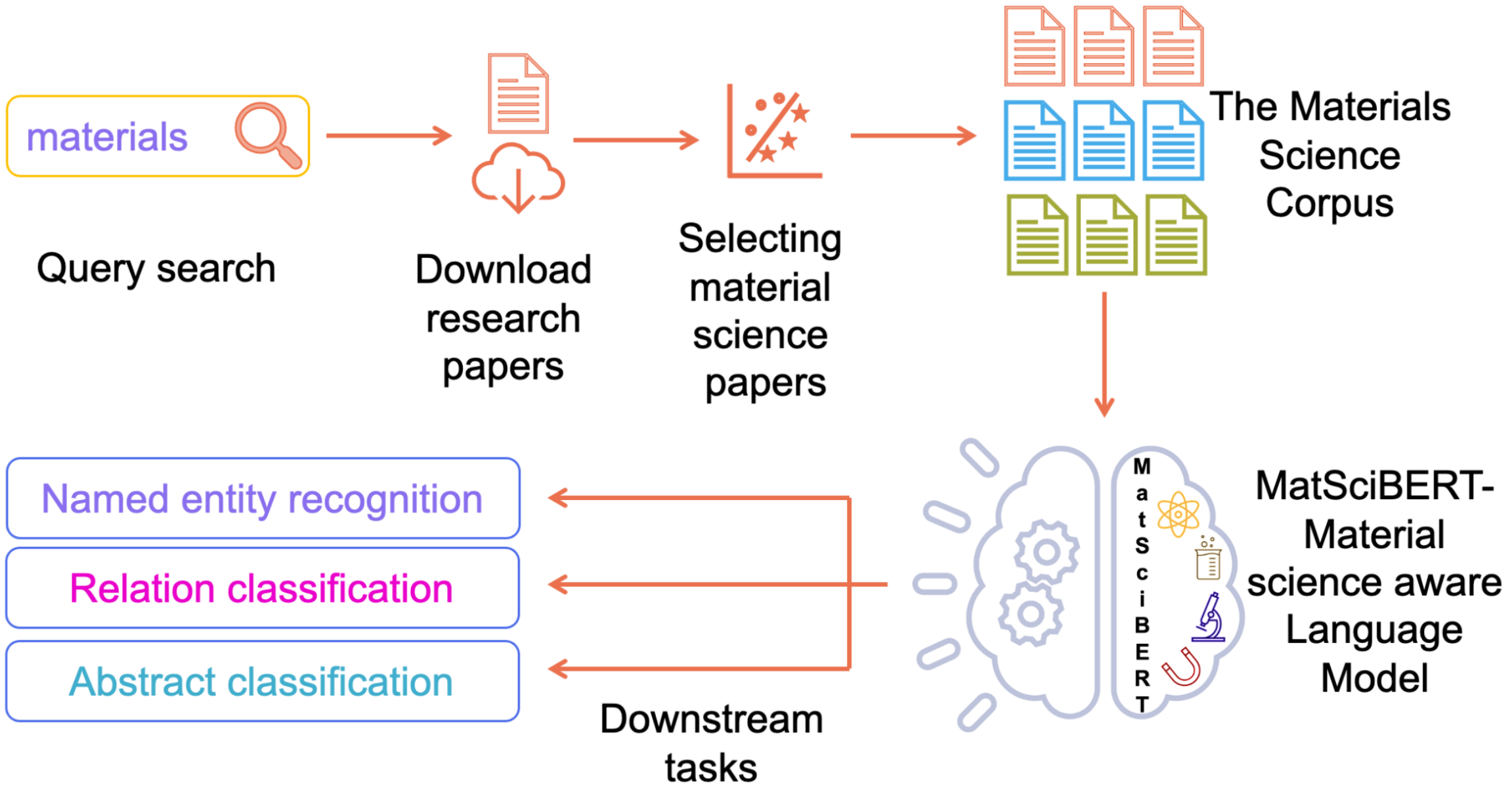

**Figure 1.** The methodology for training MatSciBERT followed by evaluation on downstream tasks.

### *2.1 Dataset collection and preparation*

In the training of an LM in a generalizable way, a considerable amount of dataset is required. For example, BERT[33] was pre-trained on BookCorpus[34] and English Wikipedia, containing a total of 3.3 billion words. SciBERT[37], an LM trained on scientific literature, was pre-trained using a corpus consisting of 82% papers from the broad biomedical domain and 18% papers from the computer science domain. However, we note that none of these LMs includes text related to the materials domain. Here, we consider materials science literature from four broad categories, namely, inorganic glasses and ceramics, metallic glasses, cement and concrete, and alloys, to cover the materials domain in a representative fashion.

The first step in retrieving the research papers is to query search from the Crossref metadata database[7]. This resulted in a list of more than 1M articles. Although Crossref gives the search results from different journals and publishers, we downloaded papers only from the Elsevier Science Direct database using their sanctioned API[8]. Note that the Elsevier API returns the research articles in XML format; hence, we wrote a custom XML parser for extracting the text. Occasionally, there were papers having only abstract and not full-text depending upon the journal and publication date. Therefore, we have included all the sections of the paper when available and abstracts otherwise. For glass science related papers, the details are given in our previous work[15]. For concrete and alloys, we first downloaded many research papers for each material category using several queries such as ‘cement’, ‘interfacial transition zone’, ‘magnesium alloy’, and ‘magnesium alloy composite materials’ to name a few.

Since all the downloaded papers did not belong to a particular class of materials, we manually annotated 500 papers based on their abstracts, whether they were relevant to the field of interest or not. Further, we finetuned SciBERT classifiers[37,49], one for each category of material, on these labelled abstracts for identifying relevant papers among the downloaded 1M articles. We consider these selected papers from each category of materials for training the language model. A detailed description of the Material Science Corpus (MSC) is given in the Results and Discussion section of the paper. Finally, we divided this corpus into training and validation, with 85% being used to train the language models and the remaining 15% to see their performance on unseen text.

Note that the texts in the scientific literature may have several symbols, including some random characters. To address these anomalies, we also performed Unicode normalization of MSC to:

A. get rid of random Unicode characters like ⬚, Ͽ, ▒, and
B. map different Unicode characters having similar meaning and appearance to either a single standard character or a sequence of standard characters.

For example, ﹪ gets mapped to %, ＞ to >, ⋙ to >>>, ⁼ and ＝ to =, ¾ to 3/4, to name a few. First, we normalized the corpus using BertNormalizer from the tokenizers library by Hugging Face[47,50]. Next, we created a list containing mappings of the Unicode characters appearing in the MSC. We mapped random characters to space so that they do not interfere during pre-training. It’s important to note that we also perform this normalization step on every dataset before passing it through MatSciBERT. This dataset was then used to pretrain the MatSciBERT.

*2.2 Pre-training of MatSciBERT*

We pre-train MatSciBERT on MSC as detailed in the last sub-section. Pre-training an LM from scratch requires significant computational power and a large dataset. To address this issue, we initialize MatSciBERT with weights from SciBERT and perform tokenization using the SciBERT uncased vocabulary. This has the additional advantage that existing models relying on SciBERT, which is pretrained on biomedical and computer science, can be interchangeably used with MatSciBERT, making both the LMs compatible with each other. Further, the vocabulary existing in the scientific literature as constructed by SciBERT may be used to represent the new words in the materials domain.

To pre-train MatSciBERT, we employ the improved training recipe suggested by Liu et al. (2019) used to train the RoBERTa[51]. Specifically, the following simple modifications were adopted for MatSciBERT pre-training: (i) Dynamic whole word masking, (ii) Removing the NSP loss from the training objective, (iii) Training on full-length sequences, (iv) Using larger batch sizes. Following these modifications, we pre-train MatSciBERT on the MSC with a maximum sequence length of 512 tokens for 10 days on 2 NVIDIA V100 32GB GPUs with a batch size of 256 sequences per GPU. We use dynamic whole-word masking as the training objective. It involves doing masking at the word level instead of masking at the WordPiece level, as discussed in the new release of BERT pre-training code[52] by Google. Each time a sequence is sampled, we randomly mask 15% of the words and let the model predict each masked WordPiece token independently. To ensure training on full length sentences, we allow input sentences to contain segments of more than one document and separate documents using the [SEP] token. We use the AdamW optimizer with $\beta_1$= 0.9, $\beta_2$= 0.98, $\varepsilon$ = $1e^{-6}$, weight decay = $1e^{-2}$ and linear decay schedule for learning rate with warmup ratio = 4.8% and peak learning rate = $1e^{-4}$. Pre-training code is written using PyTorch and Transformers library and is available at our GitHub repository for this work.

### *2.3 Downstream Tasks*
Once the LM is pre-trained, we finetune it on various supervised downstream tasks. This is done so as to adapt the model to specific tasks as well as to learn the task-specific randomly initialized weights present in the output layer. We evaluate the performance of MatSciBERT on the following three downstream NLP tasks:

(i) Named Entity Recognition
(ii) Text/Abstract Classification
(iii)Relation Classification

#### *2.3.1 Tasks Description*
(i) Named Entity Recognition (NER) involves identifying domain-specific named entities in a given sentence. Entities are encoded using the BIO scheme to account for multi-token entities[53]. Dataset for the NER task includes various sentences, with each sentence being split into multiple tokens. Gold labels are provided for each token. More formally, Let E = $\{e_1, \ldots e_k\}$ be the set of k entity types for a given dataset. If $[x_1, \ldots x_n]$ are tokens of a sentence and $[y_1, \ldots y_n]$ are labels for these tokens, then each $y_i \in L = \{B\text{-}e_1, I\text{-}e_1, \ldots B\text{-}e_k, I\text{-}e_k, O\}$.

(ii) Input for the Relation Classification[54] task consists of a sentence and an ordered pair of entity spans in that sentence. Output is a label denoting the directed relationship between the two entities. The two entity spans can be represented as $s_1 = (i, j)$ and $s_2 = (k, l)$, where i and j denote the starting and ending index of the first entity and similarly k and l denote the starting and ending index of the second entity in the input statement. Here $i \leq j$, $k \leq l$ and $(j < k \text{ } or \text{ } l < i)$. The last constraint guarantees that the two entities do not overlap with each other. The output label belongs to L, where L is a fixed set of relation types.

(iii) In the Text/Abstract Classification task, we are given an abstract of a research paper, and we have to classify whether the abstract is relevant to a given field or not.

#### *2.3.2 Datasets*

We use the following three Material Science based NER datasets to evaluate the performance of MatSciBERT against SciBERT:

1. Matscholar NER dataset[14,55] by Weston et al. (2019): This dataset is publicly available and contains 7 different entity types. Training, validation and test set consists of 440, 511 and 546 sentences, respectively. Entity types present in this dataset are inorganic material (MAT), symmetry/phase label (SPL), sample descriptor (DSC), material property (PRO), material application (APL), synthesis method (SMT) and characterization method (CMT).

2. Solid Oxide Fuel Cells – Entity Mention Extraction (SOFC) dataset[44] by Friedrich et al. (2020): This dataset consists of 45 open-access scholarly articles annotated by domain experts. Four different entity types have been annotated by the authors, namely Material, Experiment, Value and Device. There are 611, 92 and 173 sentences in the training, validation and test sets.

3. Solid Oxide Fuel Cells – Slot Filling (SOFC-Slot) dataset[44] by Friedrich et al. (2020): This is the same as the above dataset except that entity types are more fine-grained. There are 16 different entity types, namely Anode Material, Cathode Material, Conductivity, Current Density, Degradation Rate, Device, Electrolyte Material, Fuel Used, Interlayer Material, Open Circuit Voltage, Power Density, Resistance, Support Material, Time of Operation, Voltage and Working Temperature. Two additional entity types: Experiment Evoking Word and Thickness, are used for training the models.

For relation classification, we use the Materials Synthesis Procedures dataset[45] by Mysore et al. (2019). This dataset consists of 230 synthesis procedures annotated as graphs where nodes represent the participants of synthesis steps, and edges specify the relationships between the nodes. The average length of a synthesis procedure is 9 sentences, and 26 tokens are present in each sentence on average. The dataset consists of 16 relation labels. The relation labels have been divided into three categories by the authors:

a. Operation-Argument relations: Recipe target, Solvent material, Atmospheric material, Recipe precursor, Participant material, Apparatus of, Condition of
b. Non-Operation Entity relations: Descriptor of, Number of, Amount of, Apparatus-attr-of, Brand of, Core of, Property of, Type of
c. Operation-Operation relations: Next operation

The train, validation, and test set consists of 150, 30 and 50 annotated material synthesis procedures, respectively.

The dataset for classifying research papers related to glass science or not on the basis of their abstracts has been taken from Venugopal et al. (2021)[15]. The authors have manually labelled 1500 abstracts as glass and non-glass. These abstracts belong to different fields of glass science like bioactive glasses, rare earth glasses, glass ceramics, thin film studies, and optical, dielectric, and thermal properties of glasses, to name a few. We divide the abstracts into a train-validation-test split of 3:1:1.

*2.4 Modelling*

*2.4.1 Named Entity Recognition*

We use the BERT contextual output embedding of the first WordPiece of every token to classify the tokens among |L| classes. We model the NER task using three architectures: LM-Linear, LM-CRF and LM-BiLSTM-CRF. Here, LM can be replaced by any BERT-based transformer model. We take LM to be SciBERT and MatSciBERT in this work.

1. LM-Linear: The output embedding of the WordPieces are passed through a linear layer with softmax activation. We use the BERT Token Classifier implementation of transformers library[50].
2. LM-CRF: We replace the final softmax activation of the LM-Linear architecture with a CRF layer[56] so that the model can learn to label the tokens belonging to the same entity mentioned and also learn the transition scores between different entity types. We use the CRF implementation of pytorch-crf library[57].
3. LM-BiLSTM-CRF: Bidirectional Long Short Term Memory[58] is added in between the LM and CRF layer. BERT embeddings of all the WordPieces are passed through a stacked BiLSTM. The output of BiLSTM is finally fed to the CRF layer to make predictions.

*2.4.2 Relation Classification*

We use the Entity Markers-Entity Start architecture[54] proposed by Soares et al. for modelling of the relation classification task. Here, we surround the entity spans within the sentence with some special WordPieces. We wrap the first and second entities with [E1], [\E1] and [E2], [\E2] respectively. We concatenate the output embeddings of [E1] and [E2] and then pass it through a linear layer with softmax activation. We use the Standard Cross Entropy loss function for the training of the linear layer and finetuning of the language model.

*2.4.3 Text/Abstract Classification*

We use the output embedding of the CLS token to encode the entire text/abstract. We pass this embedding through a simple classifier to make predictions. We use the BERT Sentence Classifier implementation of the transformers library[50].

*2.4.4 Hyperparameters*

We use a linear decay schedule for the learning rate with a warmup ratio of 0.1. To ensure sufficient training of randomly initialized non-BERT layers, we set different learning rates for the BERT part and non-BERT part. We set the peak learning rate of the non-BERT part to 3e-4 and choose the peak learning rate of the BERT part from [2e-5, 3e-5, 5e-5], whichever results in a maximum validation score averaged across 3 seeds. We use a batch size of 16 and AdamW optimizer for all the architectures. For LM-BiLSTM-CRF architecture, we use a 2-layer stacked BiLSTM with a hidden dimension of 300 and dropout of 0.2 in between the layers. We perform finetuning for 15, 20 and 40 epochs for Matscholar, SOFC and SOFC Slot datasets, respectively, as initial experiments exhibited little or no improvement after the specified number of epochs. All the weights of any given architecture are updated during finetuning, i.e., we do not freeze any of the weights. We make the code for finetuning and different architectures publicly available. We refer readers to the code for further details about the hyperparameters.

*2.4.5 Evaluation Metrics*

We evaluate the NER task based on entity-level exact matches. We use the CoNLL evaluation script (https://github.com/spyysalo/conlleval.py) after verifying its correctness. For NER and Relation Classification tasks, we use Micro-F1 and Macro-F1 as the primary evaluation metrics. We use binary F1-score to evaluate the performance of the Text/Abstract classification task.

## 3 Results and discussion

*3.1 Dataset*

Textual datasets are an integral part of the training of an LM. There exist many general purpose corpora like BookCorpus and EnglishWikipedia[33,34], and domain specific corpora like biomedical corpus[38], and clinical database[40], to name a few. However, none of these corpora is suitable for the materials domain. Therefore, with the aim of providing a materials specific LM, we first create a corpus spanning four important material science families of inorganic glasses, metallic glasses, alloys, and cement and concrete. It should be noted that although these broad categories are mentioned, several other categories of materials, including two-dimensional materials, were also present in the corpus. Specifically, we have selected ~150K papers out of ~1M papers downloaded from the Elsevier Science Direct Database. The steps to create the corpus are provided in the *Methodology section.* The details about the number of papers and words for each family are given in Table 1.

The material science corpus developed for this work has ~285M words, which is nearly 9% of the number of words used to pre-train SciBERT (3.17B words) and BERT (3.3B words). From Table 1, one can observe that 40% of the words are from research papers related to inorganic glasses and ceramics, and 20% each from bulk metallic glasses (BMG), alloys, and cement. Note that although the number of research papers for "cement and concrete" is more than "inorganic glasses and ceramics", the latter has higher words. This is because of the presence of a greater number of full text documents retrieved associated with the latter category. The average paper length for this corpus is ~1848 words, which is two-thirds of the average paper length of 2769 words for the SciBERT corpus. The lower average paper length can be attributed to two things: (a) In general, material science papers are shorter than biomedical papers. We verified this by computing the average paper length of full text material science papers. The number came out to be 2366. (b) There are papers without full text also in our corpus. In that case, we have used the abstracts of such papers to arrive at the final corpus.

| **Corpus details** | **# (Papers)** | **# (Words)** |
|---|---|---|
| Inorganic glasses and ceramics | 42,186 | 112,560,687 |
| Bulk metallic glasses | 21,093 | 59,338,072 |
| Alloys | 21,093 | 55,683,291 |
| Cement and concrete | 69,606 | 56,936,694 |
| **Total** | **153,978** | **284,518,744** |

**Table 1.** Numbers of papers and tokens taken for training the language model

*3.2 Pre-training of MatSciBERT*

Since training from scratch is a computationally intensive task, we resort to the way BioBERT was pre-trained[36], that is, we initialize MatSciBERT weights with that of some suitable LM and then pre-train it on MSC. To determine the appropriate initial weights for MatSciBERT, we trained an uncased WordPiece[46] vocabulary based on the MSC using the tokenizers library[47]. The overlap of MSC vocabulary is 53.64% with the uncased SciBERT[37] vocabulary and 38.90% with the uncased BERT vocabulary. Because of the larger overlap with the vocabulary of SciBERT, we tokenize our corpus using the SciBERT vocabulary and initialize the MatSciBERT weights with that of SciBERT as made publicly available by Beltagy et al. (2019)[37]. The details of the pre-training procedure are provided in the *Methodology* section.

Figure 2 shows the variation of perplexity (ppl) on the validation data with the number of days for which pre-training was done. The model achieved a final perplexity of 3.112 on the validation set. The final pre-trained LM was then used for evaluation on different material science domain-specific downstream tasks, details of which are described in the subsequent sections. The performance of the LM on the downstream tasks was compared with that of SciBERT to evaluate the effectiveness of MatSciBERT to learn the materials specific information.

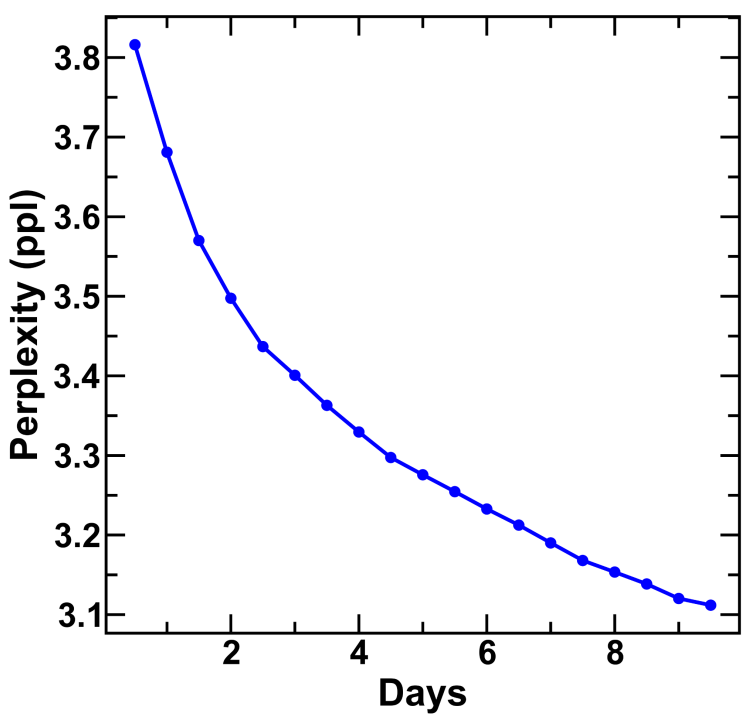

**Figure 2.** Perplexity vs number of days taken for pre-training

In order to understand the effect of pre-training on the model performance, a materials domain specific downstream task, NER on SOFC-slot, was performed on the model at regular intervals of pre-training. To this extent, the pre-trained model was finetuned on the training set of the SOFC-slot dataset. The choice of the SOFC-slot dataset was based on the fact that the dataset was comprised of fine-grained materials specific information. Thus, this dataset is appropriate to distinguish the performance of SciBERT from the materials-aware LMs. The performance of these finetuned models was evaluated on the test set. LM-CRF architecture was used for the analysis since LM-CRF consistently gives the best performance for the downstream task, as shown later in this work. Figure 4 shows the Macro-F1 averaged across 3 seeds for the SOFC-Slot test set. LM-CRF architecture varies with the number of days for which MatSciBERT was pre-trained. The increasing trend of the graph shows the importance of pre-training for longer durations.

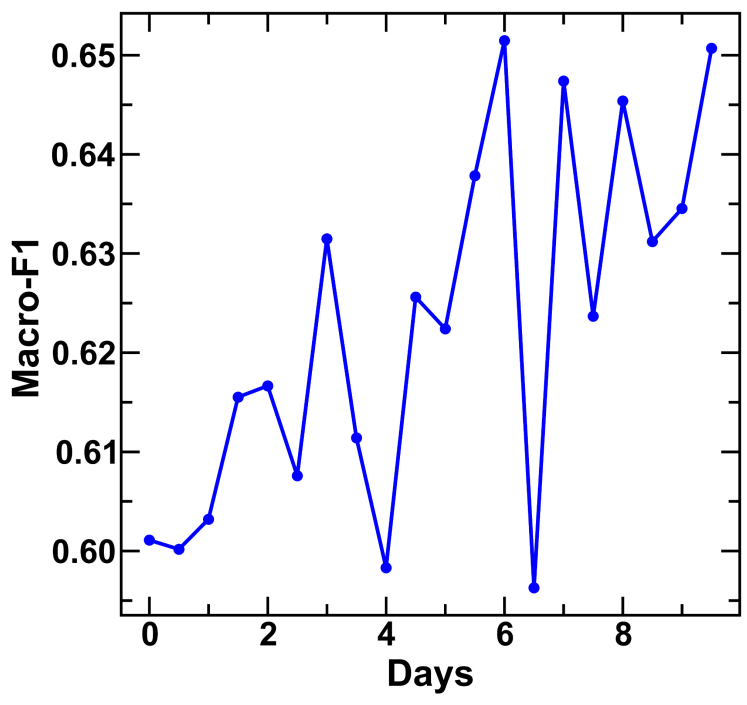

**Figure 3.** Macro-F1 (test set) vs number of days elapsed for MatSciBERT training

*3.3 Downstream tasks*

3.3.1 Named Entity Recognition

Here, we present the results on the three material science NER datasets as described in the methodology section. To the best of our knowledge, the best Macro-F1 on solid oxide fuel cells (SOFC) and SOFC-Slot datasets are 81.50% and 62.60%, respectively, as reported by Friedrich et al. (2020), who introduced the dataset[44]. We run the experiments on the exact same train-validation-test splits as done by Friedrich et al. for a fair comparison of results. Moreover, since the authors reported results averaged over 17 entities (the extra entity is "Thickness") for the SOFC-Slot dataset, we also report the results taking the "Thickness" entity into account.

| **Architecture** | **Macro - F1** | | **Micro - F1** | |
|---|---|---|---|---|
| | **LM = MatSciBERT** | **LM = SciBERT** | **LM = MatSciBERT** | **LM = SciBERT** |
| LM - Linear | 63.63% (67.40%) | 58.64% (64.58%) | 71.03% (69.64%) | 67.85% (67.58%) |
| LM - CRF | 65.00% (69.84%) | 59.07% (68.31%) | 72.06% (72.14%) | 69.68% (70.15%) |
| LM - BiLSTM-CRF | 65.92% (69.88%) | 61.68% (68.44%) | 72.72% (72.55%) | 70.66% (70.56%) |

**Table 2.** Test set results for SOFC-Slot averaged over 3 seeds and 5 cross-validation splits. Values in the parenthesis show the results on the validation set.

| **Architecture** | **Macro - F1** | | **Micro - F1** | |
|---|---|---|---|---|
| | **LM = MatSciBERT** | **LM = SciBERT** | **LM = MatSciBERT** | **LM = SciBERT** |
| LM–Linear | 81.56% (81.10%) | 79.91% (80.91%) | 83.69% (81.77%) | 81.93% (81.61%) |
| LM–CRF | 81.92% (82.53%) | 81.07% (82.04%) | 84.39% (83.22%) | 83.46% (83.03%) |
| LM–BiLSTM-CRF | 82.07% (82.20%) | 80.12% (81.92%) | 84.37% (82.93%) | 82.48% (82.78%) |

**Table 3.** Test set results for SOFC averaged over 3 seeds, and 5 cross-validation splits. Values in the parenthesis shows the results on the validation set.

Table 2 and Table 3 shows the results for the NER tasks on the SOFC-Slot and SOFC datasets, respectively, by SciBERT and MatSciBERT. We observe that LM-CRF always performs better than LM-Linear. This can be attributed to the fact that the CRF layer can model the BIO tags accurately. We obtained an improvement of ~4.2 Macro F1 and ~2.1 Micro F1 on the SOFC-Slot test set while using the LM-BiLSTM-CRF architecture. For the SOFC test dataset, MatSciBERT-BiLSTM-CRF performs better than SciBERT-BiLSTM-CRF by ~1.9 Macro F1 and ~1.9 Micro F1. Similar improvements can be seen for other architectures as well. These MatSciBERT results also surpass the current best results on SOFC-Slot and SOFC datasets by ~3.3 and ~0.6 Macro-F1, respectively.

It is worth noting that the SOFC-slot dataset consists of 17 entities and hence has more fine-grained information regarding the materials. On the other hand, SOFC-slot has only four entities representing coarse-grained information. We notice that the performance of MatSciBERT on SOFC-slot is significantly better than that of SciBERT. To further evaluate this aspect, we analyzed the F1-score of both SciBERT and MatSciBERT on all the 17 entities of the SOFC-slot data individually, as shown in Table 4. Interestingly, we observe that for all the materials related entities, namely anode material, cathode material, electrolyte material, interlayer material, and support material, MatSciBERT performs better than SciBERT. In addition, for materials related properties such as conductivity and degradation rate, MatSciBERT is able to significantly outperform SciBERT. This suggests that MatSciBERT is indeed able to capitalize on the additional information learned from the MSC to deliver better performance on complex problems specific to the materials domain.

| Entity Type | MatSciBERT F1-score | SciBERT F1-score | Entity Type | MatSciBERT F1-score | SciBERT F1-score |
|---|---|---|---|---|---|
| Anode Material | 41.72% | 38.44% | Interlayer Material | 35.26% | 28.40% |
| Cathode Material | 49.39% | 44.44% | Open Circuit Voltage | 67.39% | 66.50% |
| Conductivity | 92.10% | 89.55% | Power Density | 97.36% | 96.54% |
| Current Density | 90.89% | 92.39% | Resistance | 83.73% | 83.53% |
| Degradation Rate | 42.40% | 25.45% | Support Material | 49.63% | 47.78% |
| Device | 69.37% | 68.52% | Thickness | 77.26% | 76.20% |
| Electrolyte Material | 54.86% | 45.69% | Time of Operation | 67.98% | 67.09% |
| Fuel Used | 66.18% | 69.43% | Voltage | 69.34% | 68.27% |
| | | | Working Temperature | 90.93% | 89.66% |

**Table 4.** Comparison of entity-level F1-score for MatSciBERT and SciBERT on validation sets of SOFC-slots.

Now, we present the results for the Matscholar dataset[14] in Table 5. For the Matscholar dataset too, MatSciBERT outperforms SciBERT as well as the current best results, as can been seen in the case of LM-CRF architecture. The authors obtained a Micro-F1 of 87.09% on the validation set and 87.04% on the test set. We observe that the best model LM-CRF with MatSciBERT has Micro-F1 values of 89.33% and 87.54%, both better than the state-of-the-art.

In order to demonstrate the performance of MatSciBERT, we demonstrate an example from the validation set of the dataset. Figure 4 shows a sentence from a manuscript related to alloys. The italicized entities are colored on the basis of their true labels. From the predictions, it was observed that both MatSciBERT and SciBERT misclassified the phrase "dislocation cell-like structure" as "PRO", which was labelled as "O" or "Other" in the dataset. Further, "ND" and "TD" were labelled as "PRO" by SciBERT but MatSciBERT was able to correctly identify these entities as "O". Overall, we observe that MatSciBERT is able to identify the context in the text related to materials, from which the entities are correctly tagged. Similarly, In Figure 5, we show the results obtained using SciBERT, where it has labelled "*YSZ*" (also called yttria-stabilized zirconia) as an electrolyte material that has a true label of "B-interlayer_material". However, MatSciBERT was able to correctly identify the label of *"YSZ"*. Note that these are some arbitrary examples selected to demonstrate the performances of both SciBERT and MatSciBERT and are not necessarily representative in nature.

**Ground truth and model predictions**

*By 86% cold rolling, acicular α' martensite microstructures change into extremely refined* dislocation cell-like structure *with an average size of 60 nm, accompanied with the development of cold rolling texture in which the basal plane normal is tilted from the plate normal direction (*ND*) toward transverse direction (*TD*) at angles of ±49 deg. for Ti-8% V alloy and ±46 deg. for (Ti-8 mass% V)-4 mass% Sn alloy.*

**Labels**

Synthesis method (SMT), Symmetry/phase label (SPL), Property (PRO), Material (MAT), Descriptor (DSC)

**Figure 4.** Visualising results on the Matscholar NER dataset. The italicized colored text correspond to true labels. The underlined colored text represents the mistakes made by SciBERT. The normal font represents the mistakes made by MatSciBERT.

| Architecture | Micro - F1 | | Macro - F1 | |
|---|---|---|---|---|
| | LM = MatSciBERT | LM = SciBERT | LM = MatSciBERT | LM = SciBERT |
| LM - Linear | 86.23% (88.26%) | 84.81% (87.78%) | 85.56% (86.63%) | 83.80% (86.05%) |
| LM - CRF | 87.54% (89.33%) | 86.31% (88.77%) | 86.30% (88.75%) | 85.04% (88.07%) |
| LM - BiLSTM-CRF | 86.84% (89.27%) | 86.38% (88.71%) | 85.66% (88.87%) | 85.66% (87.66%) |

**Table 5.** Test set results for Matscholar averaged over 3 seeds. Values in the parenthesis shows the results on the validation set.

*Prediction of SciBERT:*

While the peak power density of the cell ( cell 3 ) with an YSZ (B-electrolyte_material)blocking layer reached approximately 35 mW / cm2 , that of the single - layered GDC - based cell ( cell 1) showed a much lesser power density below approximately 0.01 mW / cm2 , as shown in Figure 5a,b.

**Figure 5.** Visualising prediction on SOFC slot dataset.

3.3.2 Text/Abstract Classification

Here, we consider the ability of LMs to classify a manuscript into glass vs non-glass topics based on an in-house dataset[15]. This is a binary classification problem, with the input being the abstract of a manuscript. Table 6 shows the comparison of F1-scores achieved by MatSciBERT and SciBERT. It can be clearly seen that MatSciBERT outperforms SciBERT by ~2.5 F1-score on both validation and test sets.

| | **MatSciBERT** | **SciBERT** |
|---|---|---|
| F1-Score | 93.57% (93.04%) | 91.04% (90.51%) |

**Table 6.** Test set results for glass vs non-glass dataset averaged over 3 seeds. Values in the parenthesis represent the results on the validation set.

3.3.3 Relation Classification

Table 7 shows the results for the relation classification task performed on the Materials Synthesis Procedures dataset[45]. MatSciBERT obtains minor improvement over SciBERT for both the metrics. To the best of our knowledge, this dataset has been evaluated in only one research article[48]. However, our results are not comparable with them because of two reasons:

a) They assume access to the entity types for making predictions while we do not,
b) We make predictions over 16 relation classes, while they include empty relation class as well.

Even in this task, we observe that MatSciBERT performs better than SciBERT consistently, although with a lower margin.

| | MatSciBERT | SciBERT |
|---|---|---|
| Macro-F1 | 87.87% (87.99%) | 87.22% (87.21%) |
| Micro-F1 | 91.26% (91.40%) | 91.04% (91.03%) |

**Table 7.** Test set results for Materials Synthesis Procedures dataset averaged over 3 seeds. Values in the parenthesis represent the results on the validation set.

### 3.4 Applications in Materials Domain

Now, we discuss some of the potential areas of application of MatSciBERT in materials science. These areas can range from the simple topic-based classification of research papers to discovering novel materials or novel applications for existing materials. We demonstrate some of these applications below.

**(i) Document classification:** A large number of manuscripts have been published on materials related topics, and the numbers are increasing exponentially. Identifying manuscripts related to a given topic is a challenging task. Traditionally, these tasks are carried out employing approaches such as term frequency-inverse document frequency (TFIDF) along with latent Dirichlet allocation (LDA). However, these approaches directly vectorize a word and are not context-sensitive. For instance, in the phrases “flat glass”, “glass transition temperature”, “tea glass”, the word “glass” is used in a very different sense. MatSciBERT, being based on BERT, will be able to extract the contextual meaning of the embeddings. Thus, MatSciBERT will be able to effectively classify the topics thereby enabling improved topic classification. This is evident from the binary classification results presented earlier, where we observe that the F1-score obtained was found to be significantly higher than the results obtained using simple logistic regression based on TF-IDF. This approach can be extended to a larger set of abstracts for unsupervised topic modelling, enabling improved identification of documents relevant to specific topics.

**(ii) Information extraction from images:** Images hold a large amount of information regarding the structure and properties of materials. A proxy to identify relevant images would be to go through the captions of all the images. However, each caption may contain multiple entities and identifying the relevant keywords might be a challenging task. To this extent, MatSciBERT finetuned on NER can be an extremely useful tool for extracting information from figure captions.

Here, we extracted entities from the figure captions used by Venugopal et al. (2021)[15] using MatSciBERT finetuned on the Matscholar NER dataset. Specifically, entities were extracted from ~1,10,000 image captions on topics related to inorganic glasses. Using MatSciBERT, we obtained 87318 entities as DSC (sample descriptor), 10633 entities under APL (application), 145324 as MAT (inorganic material), 76898 as PRO (material property), 73241 as CMT (characterization method), 33426 as SMT (synthesis method), and 2676 as SPL (symmetry/phase label). Figure 6 shows the top 10 extracted entities under the seven categories proposed in the Matscholar dataset. The top entities associated with each of the categories are coating (application), XRD (characterization), glass (sample descriptor, inorganic material), composition (material property), heat (synthesis method), and hexagonal (symmetry/phase). Further details associated with each category can also be obtained from these named entities. It should be noted that each caption may be associated with more than one entity. These entities can then be used to obtain relevant images for specific queries such as "XRD measurements of glasses used for coating" or "emission spectra of doped glasses" or "SEM images of bioglasses with Ag", to name a few.

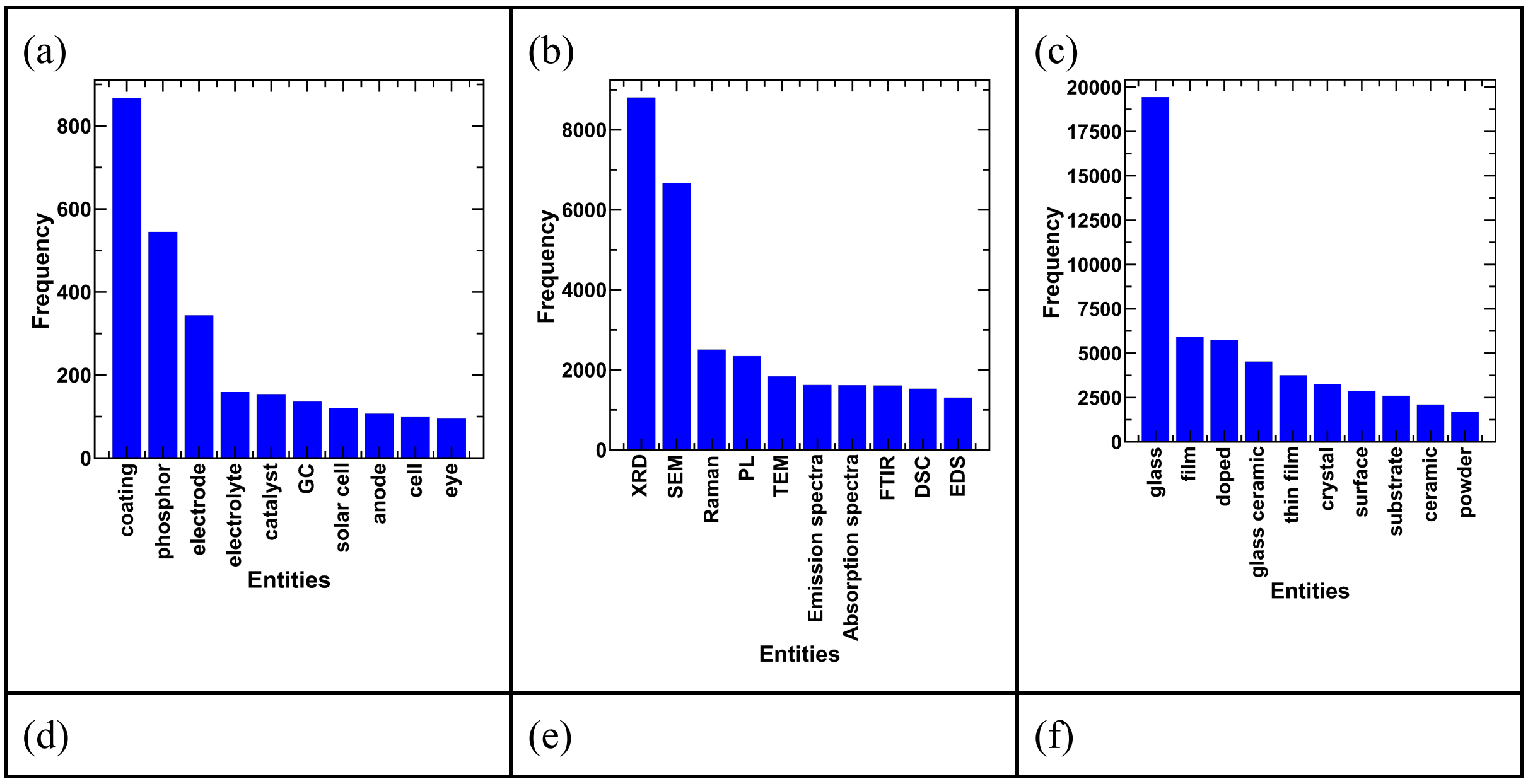

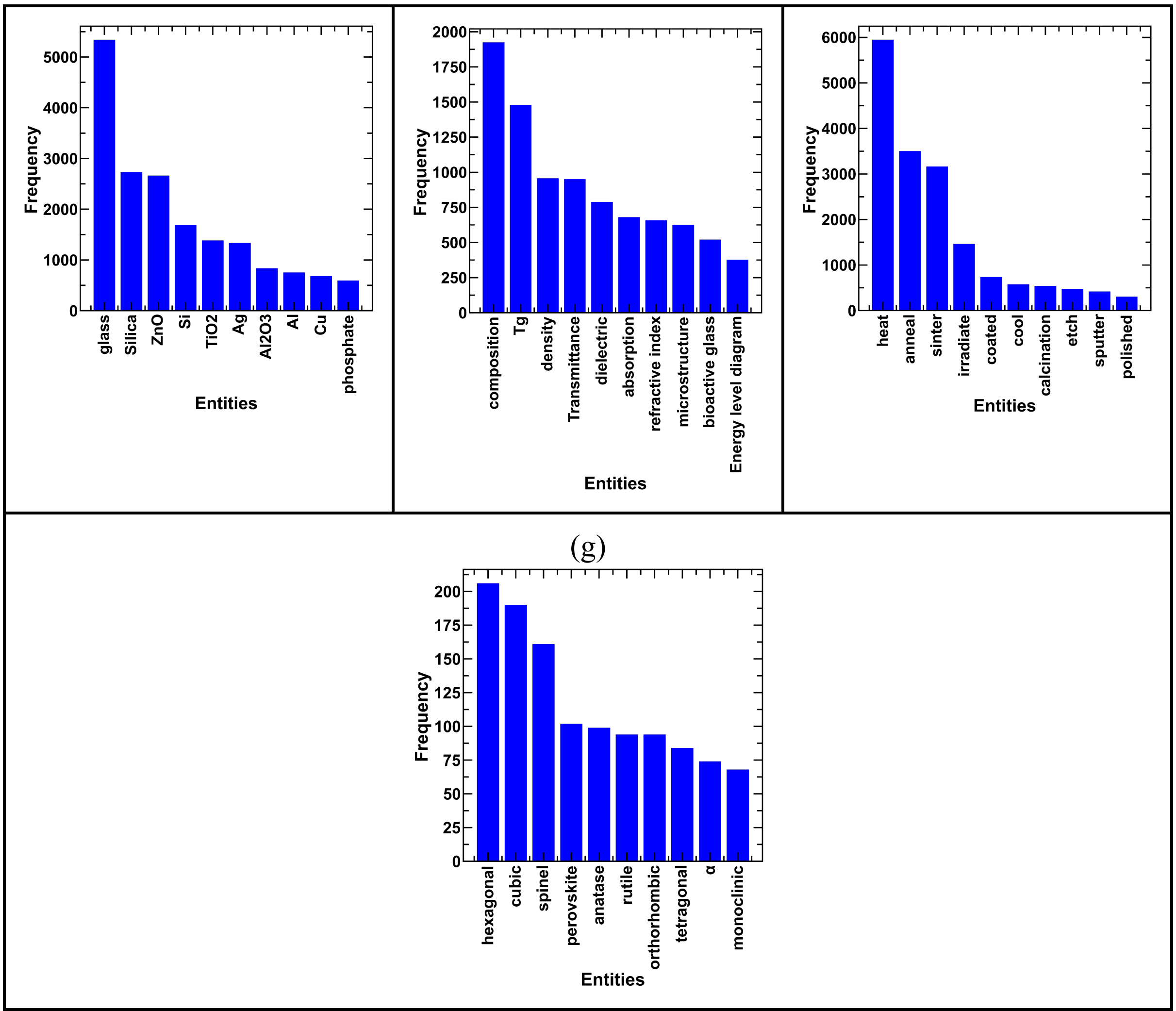

**Figure 6.** Bar graph of top-10 entities under the following categories (a) APL=Application, (b) CMT=Characterization method, (c) DSC=Sample descriptor, (d) MAT=Inorganic material, (e) PRO=Material Property, (f) SMT=Synthesis method, and (g) SPL=Symmetry/phase label

Further, Table 8 shows some of the selected captions from the image captions along with the corresponding manual annotation by Venugopal et al. (2021)[15]. These tags assigned to each caption in Venugopal et al. (2021)[15] were carried out by human experts. Note that only one word was assigned per image caption in the previous. Using the MatSciBERT NER model, we show that multiple entities are extracted for the selected 5 captions. This illustrates the large amount of information that can be captured using the LM proposed in this work.

| **Caption** | **Label by Venugopal et al. (2021)[15]** |
|---|---|
| The comparison of XRD patterns of glass ceramic heat treated at 725¬∞C for 5h and rhombohedral Ba4Gd3F17. The superstructure reflections are marked with ,óã. Inset: enlarged sections of XRD patterns. | Reflection |
| HRTEM image and the corresponding FFT pattern taken from as-deposited sample B(80/1) (a) and annealed sample D(30/1) (b); identifying rutile TiO2 crystal grains. | FFT |
| The illustrative schemes: a) The bonding of hexagonal ZnO nanocrystals to the glass surface. b) The structure of multi-layers coatings. | Crystal |
| (a) XRD patterns of the glass-ceramics sintered different holding times; (b) intensity of μ- and α-cordierite peaks count at (101) and (110) plane respectively as a function of sintering holding time. | XRD |
| Photoluminescence spectra of PbBr-based layered perovskites with an organic layer of naphthalene-linked ammonium molecules. Profiles: (a) 1; (b) 2; (c) 3; (d) 4; (e) 5. | Luminescence |
| Table 8. Comparing the results of MatSciBERT NER with manually assigned labels[15]. Application (APL), Characterization method (CMT), Descriptor (DSC), Material (MAT), Property (PRO), Synthesis method (SMT), Symmetry/phase label (SPL) | |

**(iii) Other applications such as relation classification and question answering:** MatSciBERT can also be applied for addressing several other issues such as relation classification and question answering. The relation classification task demonstrated in the present manuscript can provide key information regarding several aspects in materials science which are followed in a sequence. These would include synthesis and testing protocols, and measurement sequences. This information can be further used to discover an optimal pathway for material synthesis or a new pathway. In addition, such approaches can also be used to obtain the effect of different testing and environmental conditions, along with the relevant parameters, on the measured property of materials. This could be especially important for those properties such as hardness or fracture toughness, which are highly sensitive to sample preparation protocols, testing conditions, and the equipment used. Thus, the LM can enable the extraction of information regarding synthesis and testing conditions that are otherwise buried in the text. Similarly, it can also enable question-answering when trained on a dataset.

At this juncture, it is worth noting there are very few annotated datasets available for the material corpus. This is in contrast to the biomedical corpus, where several annotated datasets are available for different downstream tasks such as relation extraction, question-answering and NER. While the development of materials science specific language model can significantly accelerate the NLP-related applications in materials, the development of annotated datasets is equally important for accelerating materials discovery.

## 4 Conclusion

Altogether, we developed a materials-aware language model, namely, MatSciBERT, that is trained a materials science corpora of journals. The LM, trained from the initial weights of SciBERT, exploits the knowledge on computer science and biomedical corpora (on which the original SciBERT was pretrained) along with the additional information on materials science. We test the performance of MatSciBERT on several downstream tasks such as document classification, NER, and relation classification. We demonstrate that MatSciBERT exhibits superior performance on all the datasets tested in comparison to SciBERT. Finally, we discuss some of the applications through which MatSciBERT can enable accelerated information extraction from the materials science text corpora.